%% file: acl_latex.tex
\documentclass[11pt]{article}

\usepackage[final]{acl}
\usepackage{url}
\usepackage{graphicx}
\usepackage{subcaption} 
\usepackage{multirow}
\usepackage{times}
\usepackage{latexsym}
\usepackage{amsmath}
\usepackage{caption}
\usepackage[T1]{fontenc}
\usepackage{algorithm}
\usepackage{algpseudocode}

\usepackage[utf8]{inputenc}
\usepackage{color, xspace}

\usepackage{microtype}

\usepackage{inconsolata}

\usepackage{graphicx}

\title{AdaSwitch: Balancing Exploration and Guidance in Knowledge Distillation via Adaptive Switching}

\author{
  \textbf{Jingyu Peng}$^{\ddagger\mathsection\dagger}$, 
  \textbf{Maolin Wang}$^{\mathsection}$, 
  \textbf{Hengyi Cai}$^{\dagger}$, 
  \textbf{Yuchen Li}$^{\dagger}$, 
  \textbf{Kai Zhang}$^{\ddagger}  \footnotemark[1]$, \\
  \textbf{Shuaiqiang Wang}$^{\dagger}$, 
  \textbf{Dawei Yin}$^{\dagger}$, 
  \textbf{Xiangyu Zhao}$^{\mathsection} \footnotemark[1]$
  \\
  $^{\ddagger}$ University of Science and Technology of China, \\
  $^{\mathsection}$ City University of Hong Kong, $^{\dagger}$ Baidu Inc. \\
  \texttt{jypeng28@mail.ustc.edu.cn}, \texttt{morin.wang@my.cityu.edu.hk}
}

\begin{document}
\maketitle
\begin{abstract}
Small language models (SLMs) are crucial for applications with strict latency and computational constraints, yet achieving high performance remains challenging. 
Knowledge distillation (KD) can transfer capabilities from large teacher models, 
but existing methods face a dilemma: off-policy distillation provides high-quality supervision but suffers from exposure bias (training–inference mismatch), while on-policy approaches ensure consistency but are limited by the low quality of student-generated outputs.
To address these issues, we propose AdaSwitch, a novel approach that dynamically combines on-policy and off-policy generation via an adaptive switching mechanism. 
AdaSwitch allows the student to explore its predictions within its capability and selectively integrates teacher guidance only when divergence exceeds a context-aware threshold. 
This paradigm preserves generation consistency while ensuring high-quality supervision. 
Experiments on three datasets demonstrate that AdaSwitch consistently improves accuracy and reasoning capability with moderate overhead. 
\end{abstract}

\input{1Introduction}
\input{2Method}

\input{3Experiments}

\input{4RelatedWork}

\input{5Conclusion}
\input{6Limitations}
\bibliography{custom}

\input{7Appendix}

\end{document}

%% file: 1Introduction.tex
\section{Introduction}

With the rapid development of large language models (LLMs), scaling laws suggest that increasing parameters generally leads to better performance. 
However, in latency-sensitive scenarios such as real-time search and recommendation systems, deploying massive LLMs is often computationally prohibitive~\cite{li2025towards,deng2025onerec}. 
Therefore, developing Small Language Models (SLMs) that retain strong performance while minimizing inference costs is of vital importance.

\begin{figure}[t]
	\centering
    \includegraphics[width= 0.99\linewidth]{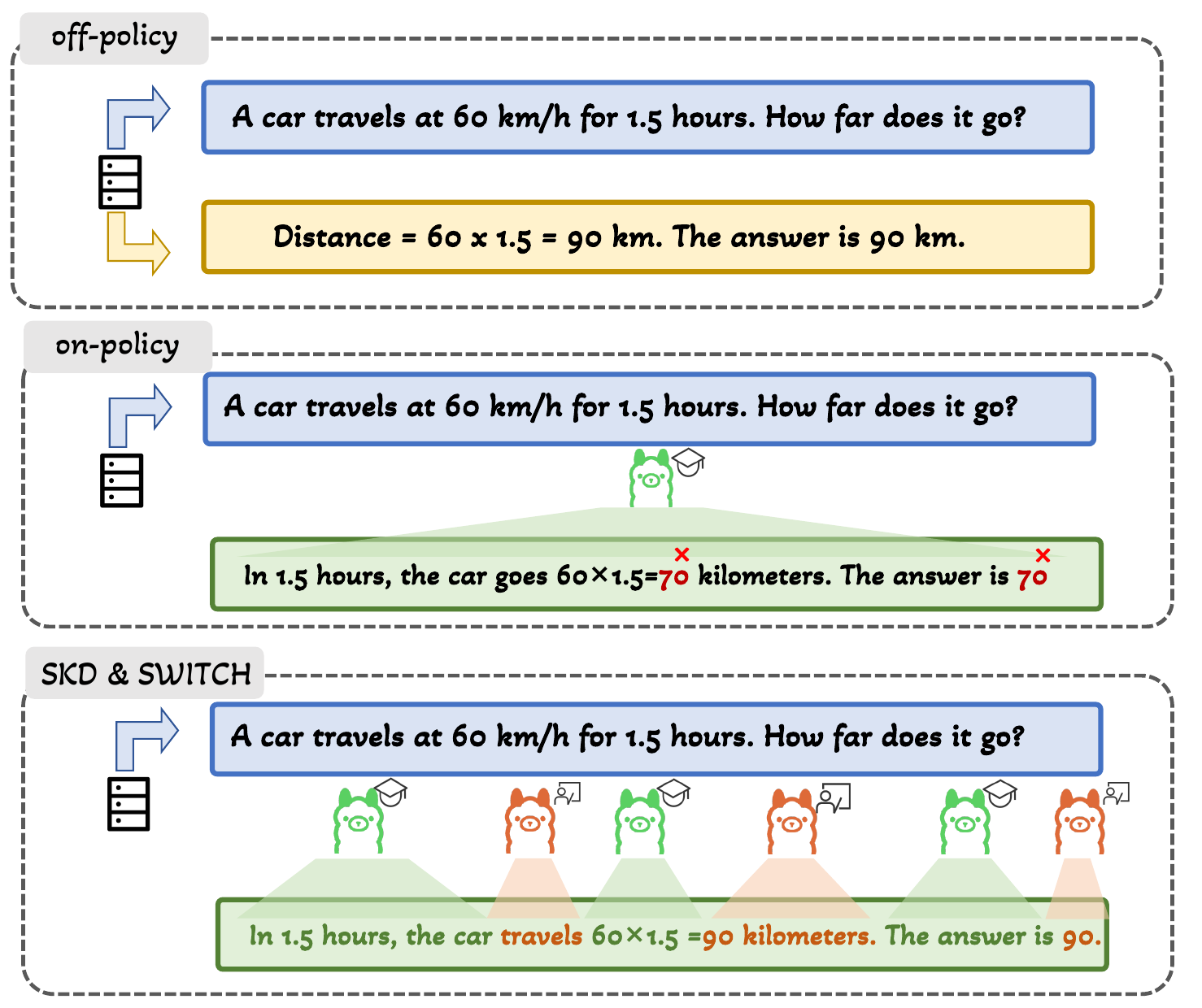}
    
    \caption{An example illustrating the sequence generation process for off-policy, on-policy, SKD, and SWITCH methods. \includegraphics[height=1.0em]{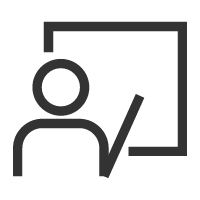} denotes the teacher, and \includegraphics[height=1.0em]{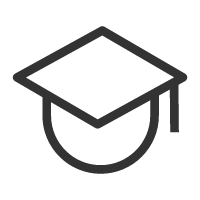} represents the student.}
    \vspace{-0.5cm}
    \label{fig:for}

\end{figure}


Obtaining high-quality SLMs is often achieved by transferring knowledge from large teacher models via Knowledge Distillation (KD)~\cite{DBLP:journals/corr/HintonVD15, xu2024survey, wang2025put, sanh2019distilbert}. 
Current KD paradigms generally fall into two categories depending on the source of training data~\cite{xu2024survey}: 
Off-policy methods (e.g., MiniLLM ~\cite{gu2024minillm}, SeqKD ~\cite{kim2016sequence}) utilize teacher-generated sequences or ground-truth data, providing rich supervision signals but suffering from training-inference mismatch, as the student never learns to recover from its own errors. 
Conversely, on-policy methods (e.g., GKD ~\cite{agarwal2024policy}) train on student-generated sequences, aligning training with inference but often collapsing due to the low quality of the student's initial outputs, especially in the early stages of distillation.
Effectively combining the strengths of both paradigms—high-quality supervision and distribution consistency—remains a key challenge.

\begin{figure*}[t]
\centering
\includegraphics[width=0.95\linewidth]{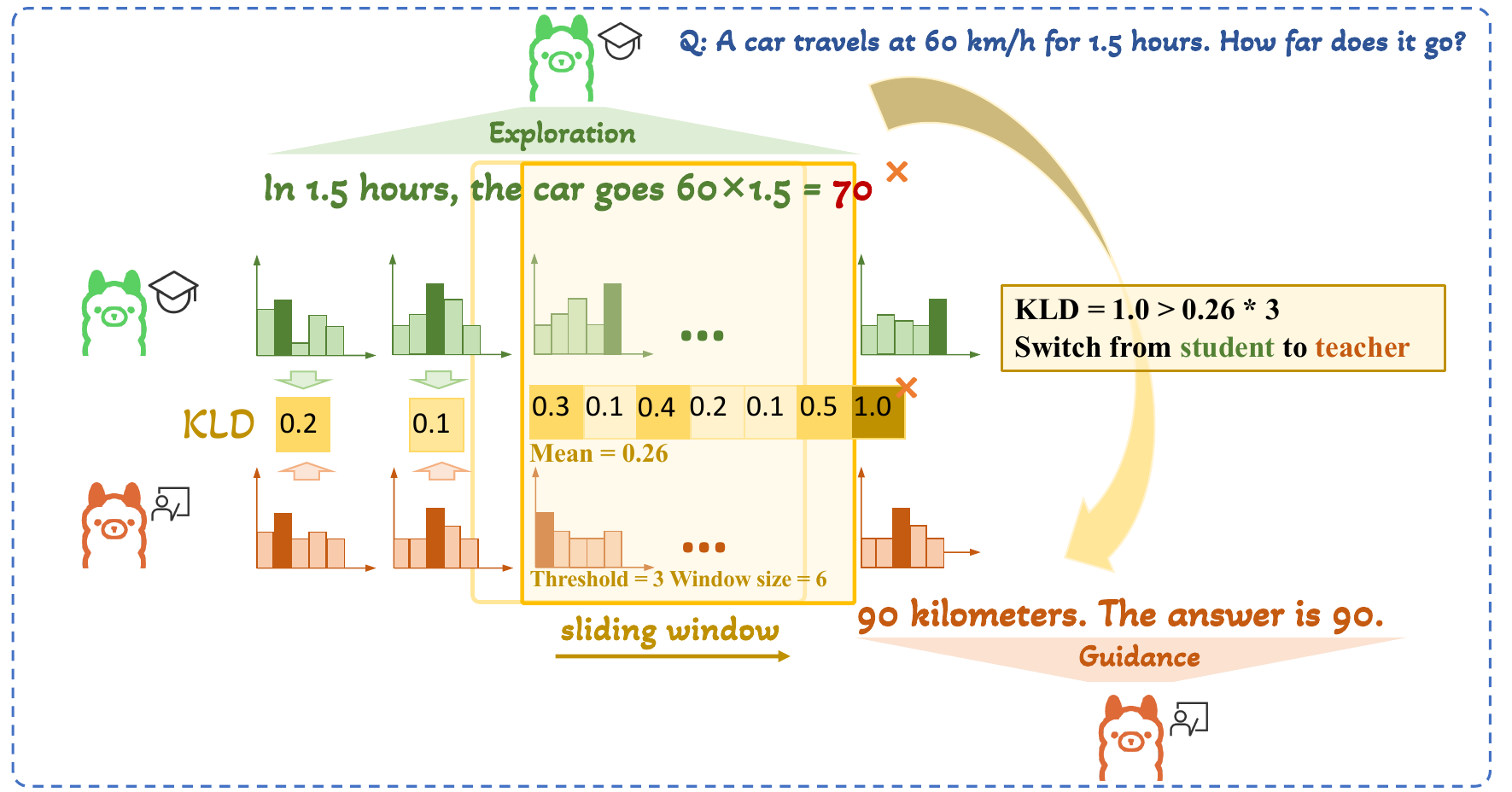}
\caption{Overview of the proposed AdaSwitch approach. When the divergence between the student and teacher logits for the next token exceeds $ \tau = K \cdot \bar{d}_i$, AdaSwitch switches to the teacher to generate the remaining tokens. \includegraphics[height=1.0em]{figure/laoshi.png} denotes the teacher, and \includegraphics[height=1.0em]{figure/xuesheng.png} represents the student.}
\label{fig:method}
\vspace{-0.5cm}
\end{figure*}

Recent attempts, such as SKD~\cite{xuspeculative} and SWITCH~\cite{koo-etal-2025-switch}, propose mixing student and teacher tokens within a single sequence. 
However, these methods typically employ frequent, bidirectional switching strategies, where control is returned to the student immediately after a teacher intervention. 
We argue that such fragmentation disrupts the semantic coherence of the generation process. 
When a model is forced to constantly toggle between the student's and the teacher's latent states, it may generate disjointed sequences (or "chimeras") that satisfy neither distribution perfectly, as validated in Section \ref{sec:fre}.
Furthermore, existing static or rigid thresholding mechanisms fail to adapt to the varying difficulty of different samples, limiting their effectiveness on complex reasoning tasks.

To address these limitations, we propose \textbf{AdaSwitch}, a context-aware sequence generation framework that dynamically alternates between on-policy exploration and off-policy guidance at the token level. 
AdaSwitch adopts an adaptive one-time switching mechanism: the student initially generates tokens autonomously (Exploration Stage) to maintain training-inference consistency.
Once the student's divergence from the teacher exceeds a dynamic threshold derived from recent generation history, the teacher takes over to generate the remaining sequence (Guidance Stage), ensuring high-quality outputs. 
This design prevents logic fragmentation caused by frequent switching and ensures that the student receives a high-quality correction exactly when its capability limit is reached.

We evaluate AdaSwitch on three datasets using three distinct teacher–student model pairs. Experimental results reveal that AdaSwitch consistently improves performance in most scenarios, highlighting its robustness and versatility across tasks. In terms of efficiency, AdaSwitch maintains an acceptable computational overhead, averaging only 1.3$\times$ that of the pure on-policy method, while achieving a 10\% reduction compared with SKD and SWITCH. This modest overhead increase ensures that AdaSwitch remains highly practical for real-world applications.

%% file: 2Method.tex
\begin{algorithm*}[t]

\caption{AdaSwitch Knowledge Distillation for LLMs}
\label{alg:ada}
\begin{algorithmic}[1]
\small
\Require Student LLM $M_s$, Teacher LLM $M_t$, Prompt dataset $\{x_n\}_{n=1}^N$, Decoding length $\alpha$, Sliding window length $L$, Divergence threshold multiplier $K$, Divergence metric $D$

\For{$n = 1$ to $N$}
    \State Initialize empty list of recent divergences $D_{\text{window}}$
    \State $use\_teacher \gets \text{False}$
    
    \For{$i = 1$ to $\alpha$}
        \If{$use\_teacher$}
            \State Sample $y_i \sim M_t(\cdot \mid y_{<i}, x_n)$ \Comment{teacher continues generating}
        \Else
            \State Sample $y_i \sim M_s(\cdot \mid y_{<i}, x_n)$ \Comment{student generates token}
            \State Compute divergence $d_i = D\big(M_t(\cdot \mid y_{<i}, x_n) \,\|\, M_s(\cdot \mid y_{<i}, x_n)\big)$
            
            \If{$i > L$}
                \State Compute sliding window average $\bar{d}_{i-1} = \frac{1}{L} \sum_{j=i-L}^{i-1} d_j$
                \If{$d_i > K \cdot \bar{d}_{i-1}$}
                    \State $use\_teacher \gets \text{True}$ \Comment{switch to teacher for remaining tokens}
                    \State Sample $y_i \sim M_t(\cdot \mid y_{<i}, x_n)$
                \EndIf
            \EndIf
            \State Append $d_i$ to $D_{\text{window}}$
        \EndIf
    \EndFor
    
    \State Apply gradient descent to minimize $D(M_t \Vert M_s)(y \mid x_n)$ over generated sequence
\EndFor
\end{algorithmic}
\end{algorithm*}

\section{Methodology}

\subsection{Problem Setup}

We study KD for LLMs by considering a teacher model $M_t$ and a student model $M_s$ with different capacities. Our goal is to train $M_s$ to mimic $M_t$’s behavior on a specific task $T$. The student model $M_s$ is parameterized by learnable weights $\theta_s$, while the teacher model $M_t$ has fixed parameters $\theta_t$.

For task $T$, we assume access to a collection of input prompts $x$ and associated target sequences $y$, either from ground-truth data or model-generated outputs. To quantify the difference between the teacher’s and student’s token-level predictions, we define a divergence function $D(\cdot \,\|\, \cdot)$, such as Kullback–Leibler (KL) divergence:

\begin{equation}
\label{eq:div}
d_i(M_t,M_s) = D\big(M_t(\cdot \mid y_{<i},x) \,\|\, M_s(\cdot \mid y_{<i},x)\big),
\end{equation}
where $i$ indexes the decoding step. The overall sequence-level divergence is then:

\begin{equation}
D(M_t \| M_s)(y|x) = \frac{1}{L_y} \sum_{i=1}^{L_y} d_i(x,y),
\end{equation}
with $L_y$ the length of $y$. The learning objective is to minimize $D(M_t \| M_s)(y|x)$, encouraging $M_s$ to closely approximate $M_t$.

\subsection{Target Sequence Selection}

The choice of target sequence $y$ is critical in KD. Off-policy methods use fixed targets from datasets, while on-policy methods generate $y$ from the student model itself:

\begin{equation}
y \sim M_s(\cdot \mid x; \theta_s).
\end{equation}

However, off-policy methods suffer from a mismatch between training and inference in the student model, whereas on-policy methods are limited by the low quality of student-generated sequences, leading to suboptimal performance.

\begin{table*}[t]
\begin{center}
    
\resizebox{0.9\linewidth}{!}{
\begin{tabular}{cccc|ccc|ccc}
\hline \hline
        & \multicolumn{3}{c}{Llama-3.1-3b to Llama-3.1-1b} & \multicolumn{3}{c}{Qwen-2.5-3b to Qwen-2.5-0.5b} & \multicolumn{3}{c}{Gemma-7b to Gemma-2b} \\
        & SUMM          & GSM           & GSM\_Plus        & SUMM          & GSM           & GSM\_Plus       & SUMM          & GSM           & GSM\_Plus      \\ \hline
Teacher & 37.5         & 59.6         & 38.6            & 37.4         & 65.4         & 45.4      & 57.0  & 36.9  & 33.4       \\
Student & 13.5         & 0.0             & 0.0     & 12.1     & 0.0        & 0.0          & 11.6  & 0.0  & 0.0           \\ \hline
SFT     & 32.3         & 29.0         & 12.0               & 30.3         & 31.0         & 15.5       & 31.8  & 18.9  & 11.7          \\
SeqKD   & 32.1         & 28.2          & 11.8            & 30.1         & 30.4         & 14.4         & 31.0  & 21.8  & 10.5        \\
Off-policy & 34.6         & 30.6         & 13.8            & 32.4         & 33.5         & 17.9      & 35.1  & 22.6  & 10.7           \\
On-policy & 34.3         & 32.7         & 14.3            & 32.3         & 34.1         & 20.2      & 34.1  & 26.2  & 13.8           \\
ImitKD  & 35.1         & 30.6         & 15.6            & 32.5          & 34.6         & 21.3        & 34.9  & 25.3  & 13.5         \\
SWITCH & 33.7          & 33.2        & 16.0             & 31.8          & 35.2       & 20.8          & 35.1  & 24.8   & 13.8 \\ 
SKD   & 34.9         & 32.4         & 16.9            & 32.3         & 33.0         & 20.4         & 35.0  & 26.0  & 13.9        \\  \hline

AdaSwitch    & \textbf{35.6}         & \textbf{35.0}         & \textbf{17.3}            & \textbf{32.6}          & \textbf{36.3}         & \textbf{21.6}        & \textbf{35.6}  & \textbf{26.8}  & \textbf{14.2}         \\ \hline \hline
\end{tabular}
}
\caption{Performance comparison of different Knowledge Distillation methods for Llama and Qwen on three tasks. Performance is evaluated using Rouge-L on the SUMM summarization dataset and accuracy on the GSM and GSM-Plus reasoning benchmarks. }
\label{tab:overall}
\end{center}
\vspace{-0.6cm}
\end{table*}

\subsection{AdaSwitch}

To overcome the limitations of off-policy and on-policy approaches and fully unlock the potential of knowledge distillation, we propose a novel sequence generation approach known as \textbf{AdaSwitch} as shown in Figure~\ref{fig:method}.

Adaswitch is designed to enhance the training of student models during sequence generation through dynamic guidance based on the divergence between the predictions made by the student and the teacher models. The primary objective of AdaSwitch is to achieve a delicate balance between exploration and guidance: while the student model is permitted to explore early predictions, it receives corrective interventions when its outputs diverge significantly from those of the teacher, thereby ensuring high-quality supervision throughout the training process.

At each generation step $ i $, we compute the divergence $ d_{i}(M_t,M_s) $ between the next-token predictions from the teacher and the student, as defined in Equation \ref{eq:div}. This divergence serves as a quantitative measure of the discrepancy between the two models' predictions on the current token, allowing us to evaluate the alignment of the student with the teacher's knowledge.

To effectively assess whether the current predictions of the student model diverge from those of the teacher model, we introduce a metric that adapts to sequences characterized by varying distributions. Specifically, we maintain a sliding window of length $ L $ over the most recent divergences, computing the moving average as follows:

\begin{equation}
\bar{d_i} = \frac{1}{L} \sum_{j=i-L+1}^{i} d_j, \quad \text{for } i \geq L.
\end{equation}

This moving average, $ \bar{d_i} $, provides a context-sensitive measure of divergence, allowing the model to adjust its behavior according to recent prediction patterns. Based on this, we define a threshold $ \tau $ when generating $y_i$ as follows:

\begin{equation}
\tau = K \cdot \bar{d}_{i-1},
\end{equation}

where $ K $ is a hyperparameter that scales the threshold relative to the moving average of divergences. The student model continues generating tokens as long as its divergence from the teacher model remains below this adaptive threshold $ \tau $. If the divergence exceeds $ \tau $, the teacher continues to generate:

\begin{equation}
    y_i \sim 
    \begin{cases}
        M_s(\cdot \mid y_{<i}, x; \theta_s), & d_{i}(M_t,M_s) \le \tau,\\ 
        M_t(\cdot \mid y_{<i}, x; \theta_t), & d_{i}(M_t,M_s) > \tau. 
    \end{cases} 
\end{equation}

Once a switch to the teacher model occurs, all subsequent tokens are generated exclusively by the teacher model, eliminating the need for further divergence checks. The complete algorithm of AdaSwitch is presented in Algorithm~\ref{alg:ada}.

By integrating adaptive sliding-window thresholds and selective teacher intervention, AdaSwitch not only ensures consistency between training and inference but also maintains high-quality supervision for the student model. This approach effectively harnesses the strengths of both exploration and guidance, leading to improved model performance and robustness in diverse sequence generation tasks.

%% file: 3Experiments.tex
\begin{figure*}[t]
    \centering
    \begin{subfigure}[b]{0.32\linewidth}
        \centering
        \includegraphics[width=\linewidth]{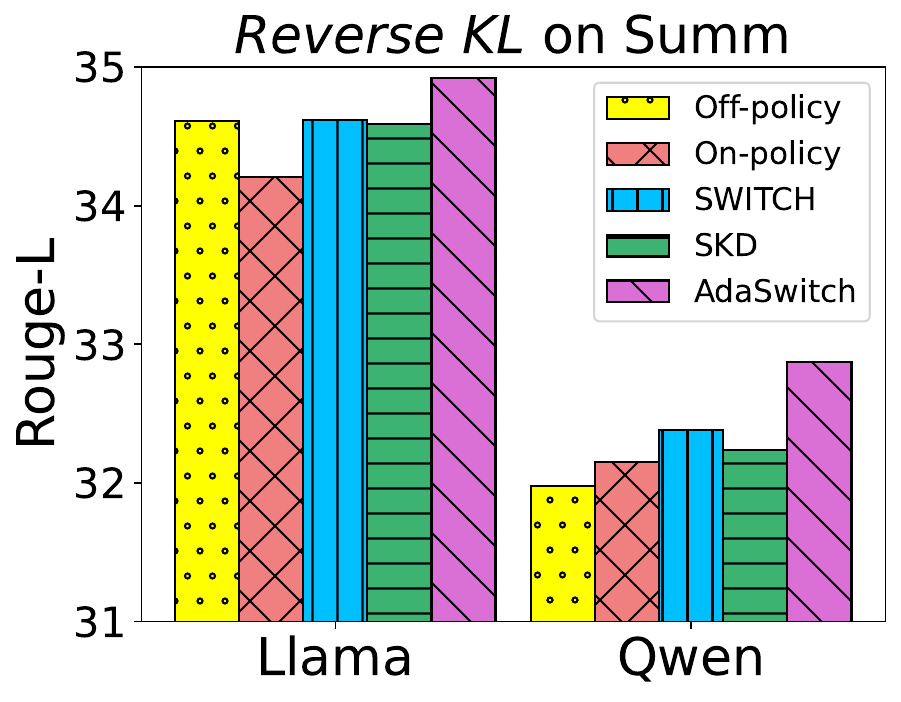}

    \end{subfigure}
    \hfill
    \begin{subfigure}[b]{0.32\linewidth}
        \centering
        \includegraphics[width=\linewidth]{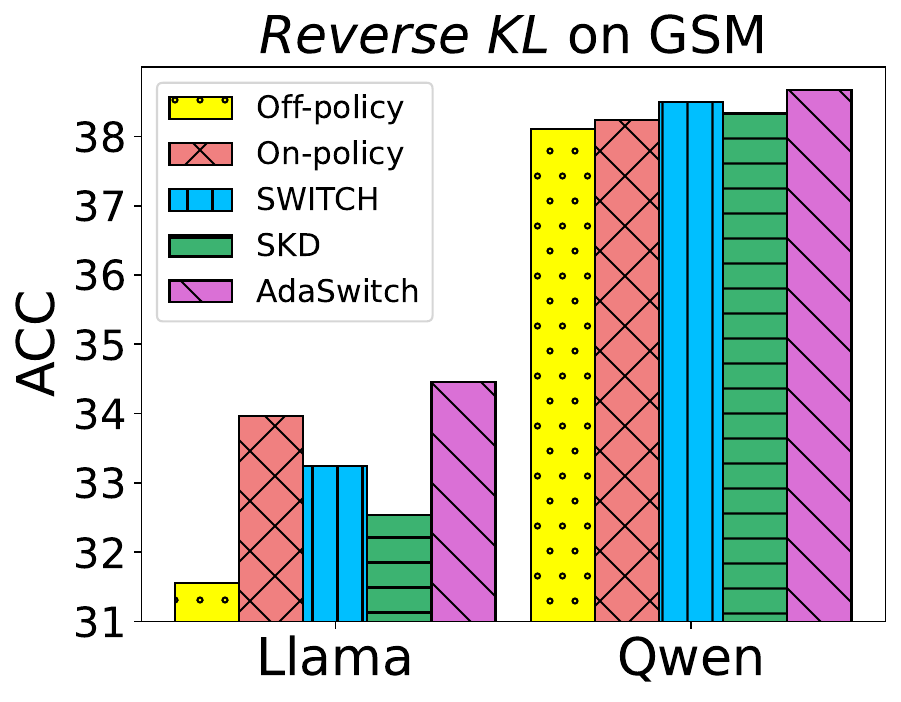}

    \end{subfigure}
    \hfill
    \begin{subfigure}[b]{0.32\linewidth}
        \centering
        \includegraphics[width=\linewidth]{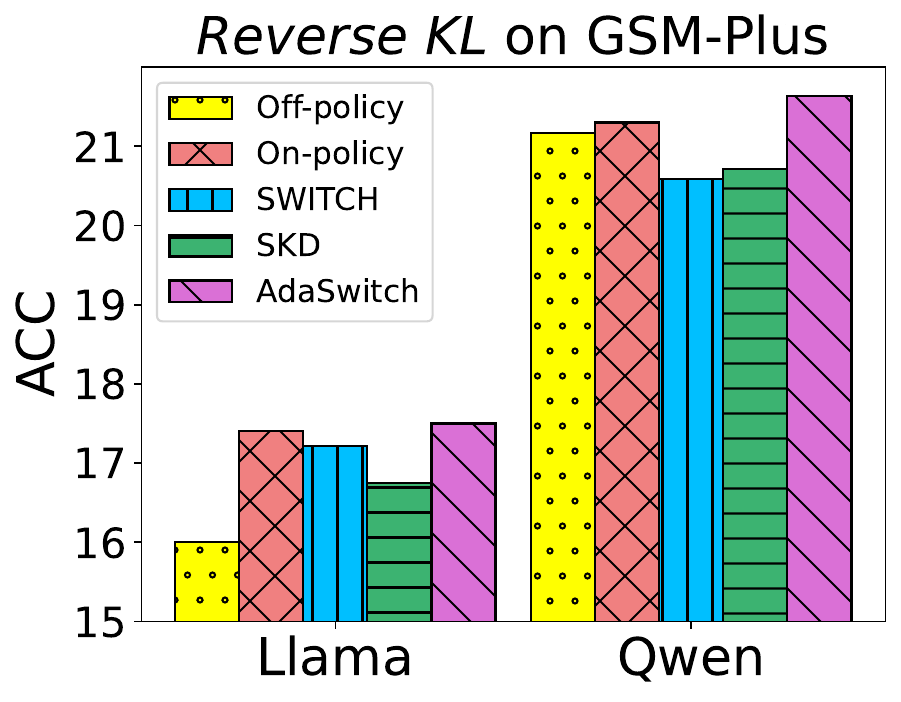}
    \end{subfigure}
    \hfill
    \begin{subfigure}[b]{0.32\linewidth}
        \centering
        \includegraphics[width=\linewidth]{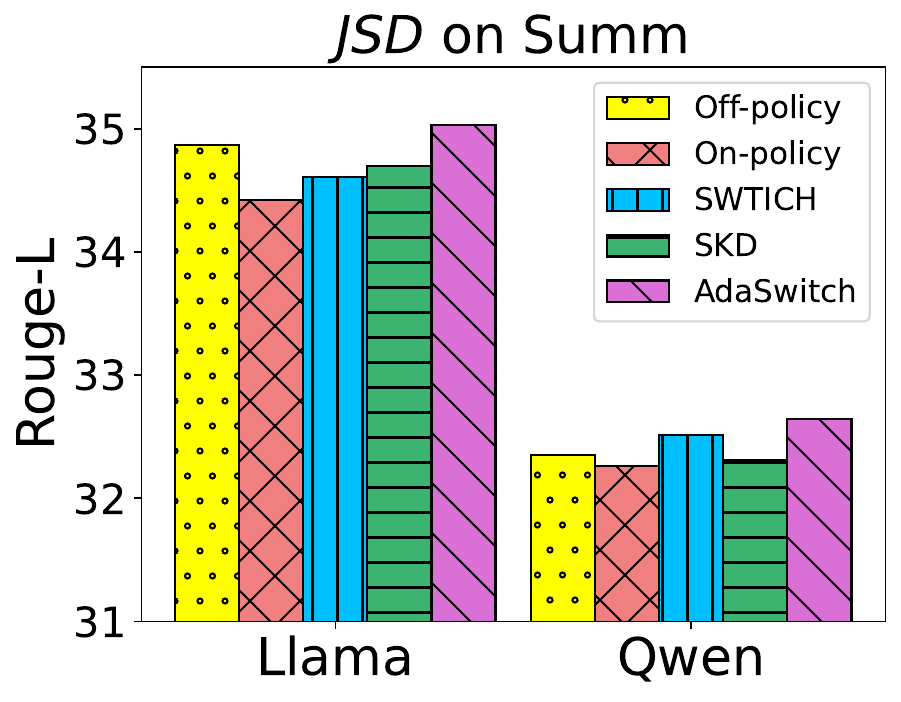}

    \end{subfigure}
    \hfill
    \begin{subfigure}[b]{0.32\linewidth}
        \centering
        \includegraphics[width=\linewidth]{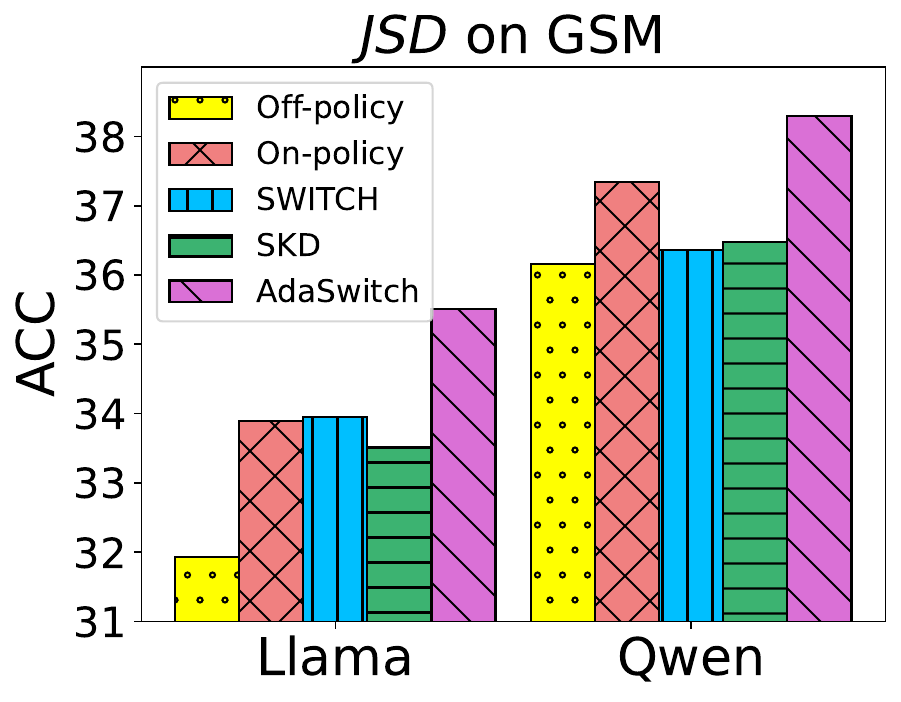}

    \end{subfigure}
    \hfill
    \begin{subfigure}[b]{0.32\linewidth}
        \centering
        \includegraphics[width=\linewidth]{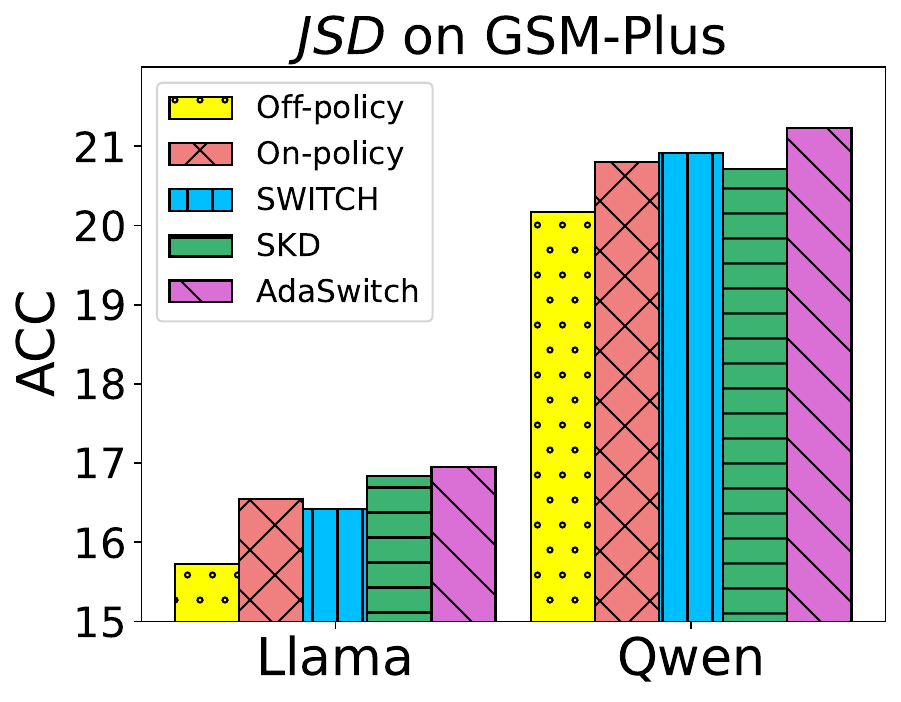}

    \end{subfigure}

    \caption{Further comparison of performance under different distance metrics on three tasks.}
    \label{fig:different_distance}
        \vspace{-0.5cm}

\end{figure*}

\section{Experiments}
\subsection{Student and Teacher Models}
We conducted experiments on three state-of-the-art LLM families, following prior work ~\cite{agarwal2024policy,koo-etal-2025-switch,xuspeculative} in selecting comparable teacher–student model scales: Qwen 2.5 \cite{team2024qwen2}, Llama 3.1 \cite{grattafiori2024llama}, and Gemma \cite{team2024gemma}. For the teacher models, we employed the SFTed Qwen-2.5-3B, Llama-3.1-3B and Gemma-7B, while the student models consisted of Qwen-2.5-0.5B, Llama-3.1-1B and Gemma-2B.

\subsection{Datasets and Experimental Setup}
To evaluate the effectiveness of our proposed AdaSwitch, we conduct experiments on two distinct types of datasets: DialogSum \cite{chen2021dialogsum} for dialogue summarization and GSM \cite{cobbe2021training} for arithmetic reasoning. For arithmetic reasoning, we use one additional dataset GSM-Plus \cite{li2024gsm} to further test the generalizability and robustness of our approach. Each task follows a two-stage procedure: we first fine-tune the teacher model on the task, under the assumption that it serves as a domain expert, and then perform knowledge distillation by randomly sampling approximately 1K examples from the task, following common practices in prior KD work. Detailed configurations for each dataset are provided in the subsequent sections. For reproducibility, all implementations adopt greedy decoding during evaluation. Detailed descriptions of the datasets, including metrics and data splits, are provided in Appendix~\ref{sec:datasets_app}.

\subsection{Hyperparameters} 

For all SFT experiments, we used a learning rate of 1e-5, a warmup ratio of 0.1, and a dropout rate of 0.1, training each model for three epochs and selecting the checkpoint with the lowest validation loss. For KD, input/output lengths were set to {1024, 256} for dialogue summarization and {128, 512} for arithmetic reasoning. Student models used a temperature of 0.5 and top-p of 0.5, while teachers used a temperature of 0.2 and top-p of 0.5, following prior work \cite{xuspeculative}. All models were trained for three epochs with a learning rate of 1e-5. For fair comparison, we set $K_{skd}=0.5$ for SKD, and fixed the sliding window length $L=10$ and threshold $K=3$. The effects of $L$ and $K$ are analyzed later.

\subsection{Baselines}
We evaluate the effectiveness of \textbf{AdaSwitch} by comparing it against six baseline methods, spanning pure on-policy and off-policy approaches, and existing sequence-level and token-level mixing methods: \textbf{Supervised FT}, \textbf{SeqKD} \cite{kim2016sequence}, \textbf{SupervisedKD} \cite{sanh2019distilbert}, \textbf{GKD} \cite{agarwal2024policy}, \textbf{ImitKD} \cite{lin2020autoregressive}, \textbf{SWITCH} \cite{koo-etal-2025-switch}, and \textbf{SKD} \cite{xuspeculative}. Due to space constraints, detailed descriptions of these baselines are provided in Appendix \ref{sec:base}.

\subsection{Overall Performance}

As shown in Table \ref{tab:overall}, on-policy and off-policy approaches exhibit complementary strengths. On-policy distillation outperforms on GSM and GSM-Plus, whereas off-policy achieves superior results on SUMM. These observations suggest that integrating both methods can be advantageous, allowing the student model to benefit from the strengths of each approach.

Mixed methods such as ImitKD, SWITCH and SKD attempt to integrate both paradigms and yield localized improvements. For instance, ImitKD performs well on SUMM, and SKD shows gains on GSM-Plus. However, these improvements are marginal and inconsistent across tasks and model pairs, reflecting the limited performance improvement of their designs.

In contrast, AdaSwitch achieves state-of-the-art results across all benchmarks, demonstrating both robustness and effectiveness. On GSM, AdaSwitch surpasses the second-best method by 7.2\% and 11.8\% when applied to Llama and Qwen, respectively. These results highlight its ability to integrate the strengths of on-policy and off-policy learning at the token level, leading to significant and consistent improvements.

For a more comprehensive comparison, we evaluate performance using JSD and reversed KLD \cite{gu2024minillm} as distance metrics, as shown in Figure~\ref{fig:different_distance}. The results indicate that, while existing mixed KD methods occasionally provide improvements, their performance is inconsistent. By comparison, although results vary slightly across different metrics, AdaSwitch consistently performs well and is compatible with different commonly used distance measures, highlighting the robustness and flexibility of AdaSwitch in diverse scenarios.

\begin{figure}[t]
    \centering
    \includegraphics[width= 0.99\linewidth]{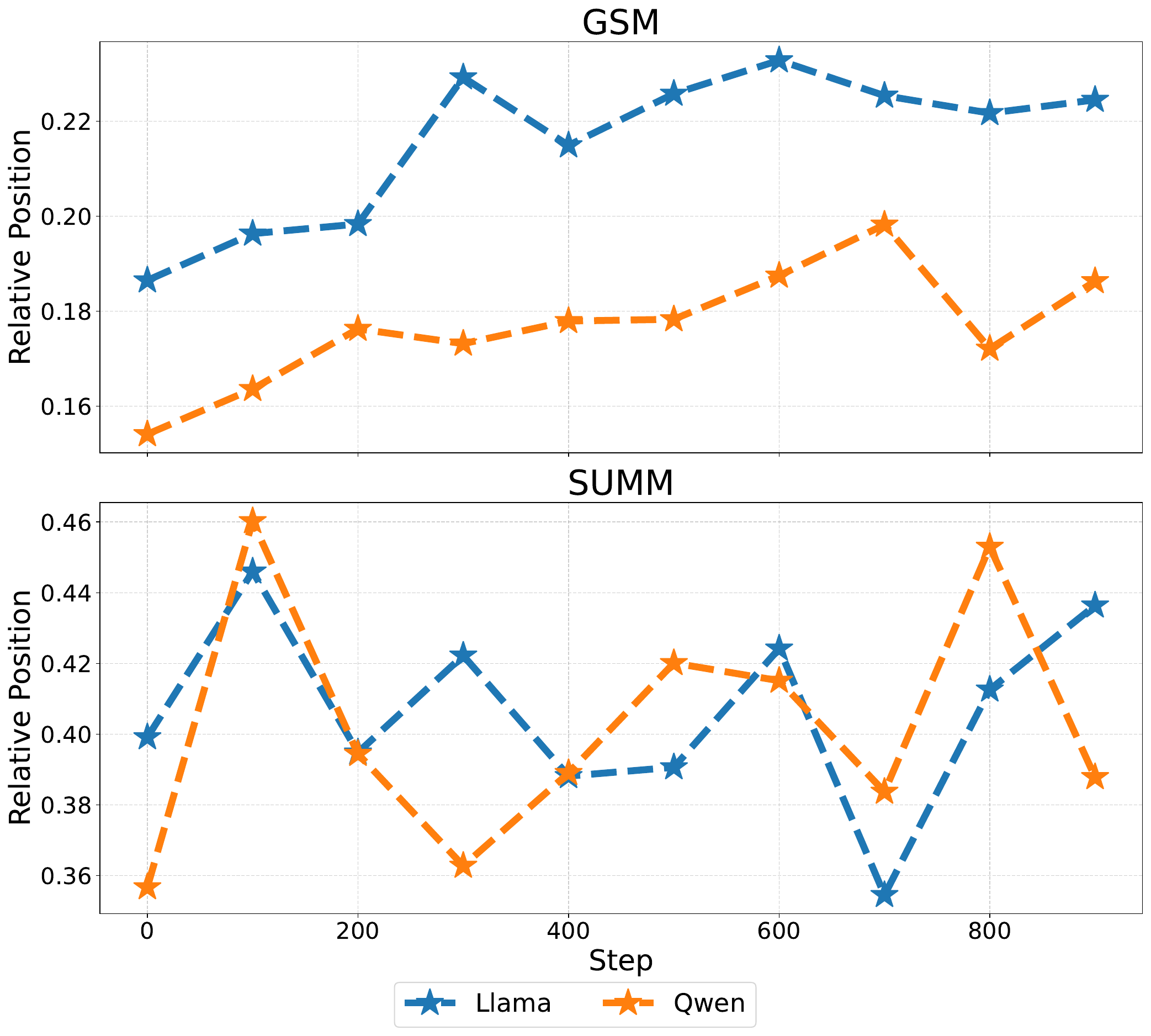}

    \caption{Analysis of the relative switching position of the distillation process}
    \label{fig:relative_position}
    \vspace{-0.5cm}
\end{figure}

\subsection{Analysis of Switching Dynamics during Distillation}
\begin{figure}[t]
    \centering
    
    \includegraphics[width= 0.9\linewidth]{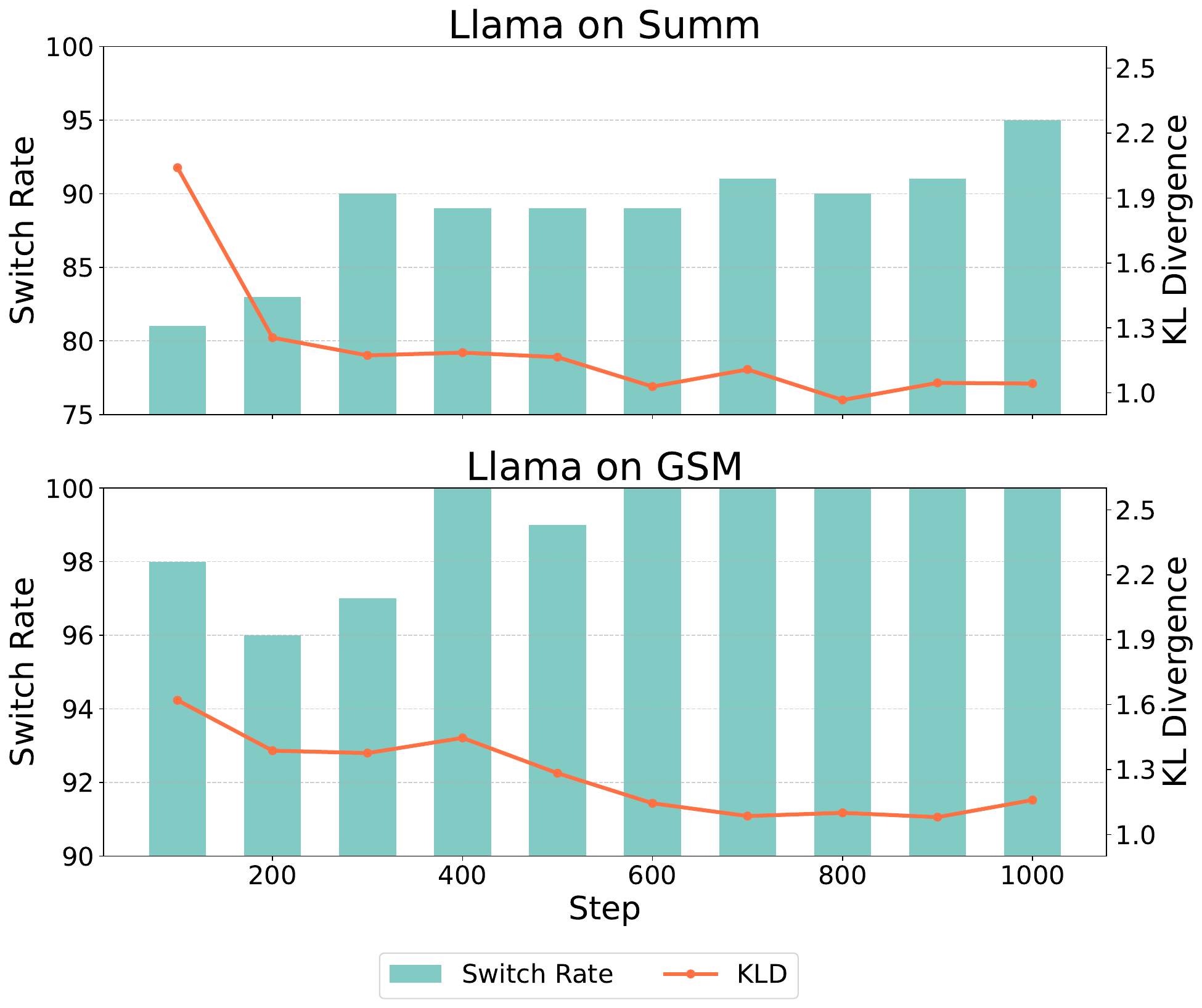}

    \caption{Analysis of the switch rate and the KLD at the switch token between the student and teacher models throughout the distillation process. Statistics were collected every 100 steps for the first 1000 steps.}
    \label{fig:switch}
    \vspace{-0.5cm}
\end{figure}

To better understand the distillation process of AdaSwitch, we analyze the switch rate, defined as the proportion of sequences that trigger a switch from exploration to guidance, along with the average KLD between the student and teacher at the switching points. Statistics were collected every 100 steps, and the results are shown in Figure~\ref{fig:switch}.

First, we observe a significant decrease in KLD during distillation across all tasks. This trend indicates that the student model gradually converges toward the teacher, consistent with intuitive expectations. Moreover, the KLD in the SUMM task is consistently lower than that in GSM, aligning with the performance difference between the two tasks and indicating that the SUMM task presents a lower reasoning complexity compared to GSM.

Switch rates also reveal task-dependent differences. On GSM, switch rates remain above 95\% throughout, whereas SUMM exhibits substantially lower rates. This suggests that more challenging tasks demand stronger reliance on teacher guidance, while easier tasks allow greater exploration. AdaSwitch thus adaptively balances exploration and guidance according to task difficulty.

Further, we observe that the switch rate increases steadily during distillation, most notably in SUMM, while remaining near 100\% in GSM. This trend reflects a gradual decline in exploration and a rise in guidance, indicating that on-policy distillation contributes less over time. Although this may seem counterintuitive, since weaker students would normally require more guidance, the pattern is explained by KLD: early in training, high divergence suppresses switching, but as the student aligns with the teacher, switching becomes more frequent. In other words, AdaSwitch begins in an exploration-dominated on-policy mode and gradually shifts to off-policy guidance as the student improves. This adaptive behavior allows AdaSwitch to dynamically balance exploration and guidance according to the student’s capability, ensuring optimal distillation across varying tasks. 

Moreover, we analyze the relative switching position of AdaSwitch during distillation. This also reflects the proportion of tokens generated by the student model. As shown in Figure~\ref{fig:relative_position}, in GSM the switching position exhibits a clear upward trend, indicating that the teacher intervenes later in the generation process. This suggests that the student model is gradually approaching the teacher’s capability. In contrast, for the SUMM task we do not observe a noticeable increase, which we attribute to the smaller performance gap between the student and teacher models in this setting.

\subsection{Early-Stage Distillation Analysis}
\begin{figure}[t]
    \centering
    \includegraphics[width= 0.95\linewidth]{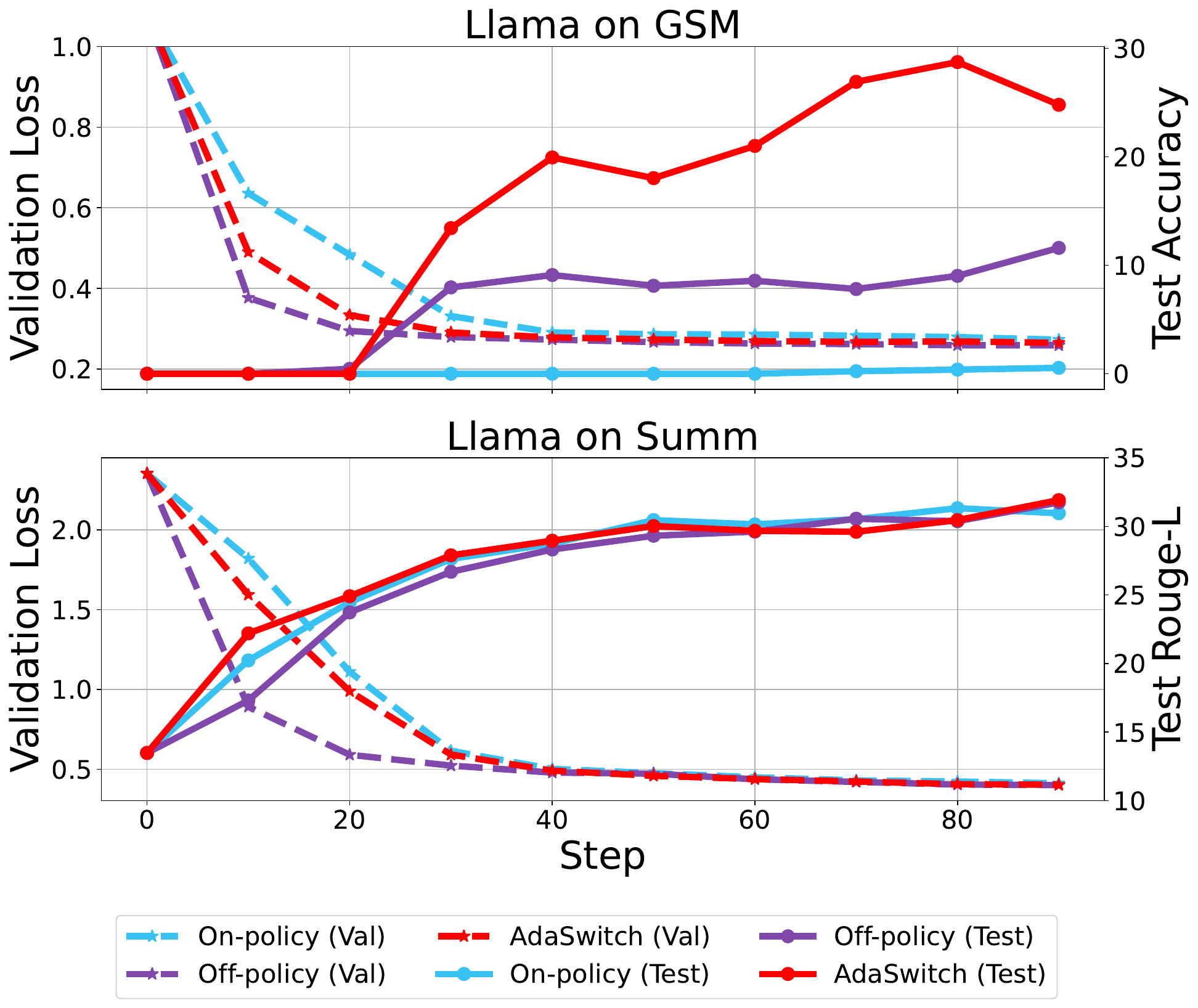}
    \caption{Performance on the validation and test sets during the early stages of distillation.}
    \label{fig:vali}
    \vspace{-0.5cm}
\end{figure}

The challenge of low-quality outputs in on-policy KD is particularly acute during the early stages of the distillation process. To investigate whether AdaSwitch addresses this issue, we present the validation loss and test performance trends for on-policy, off-policy, and AdaSwitch methods over the first 100 steps in Figure~\ref{fig:vali}.

Regarding the validation loss trends, the result indicates that AdaSwitch consistently achieves values intermediate between those of the on-policy and off-policy methods, providing evidence of an effective integration. Since validation loss is computed using the teacher model’s outputs as ground truth, the off-policy method, which directly optimizes against these outputs, attains the lowest validation loss.

Further experimental results of the performance on the test set reveal that AdaSwitch demonstrates the most rapid improvement rate, particularly on the GSM dataset. Notably, at step 30, the accuracy of AdaSwitch surpasses 10\%, whereas the off-policy and on-policy methods achieve only 7.96\% and 0\%, respectively. These results suggest that AdaSwitch, through effective token-level mixing, simultaneously ensures the quality of generated sequences and reduces the discrepancy between training and inference sequences during the early stages of distillation. The full experimental results are presented in Appendix~\ref{sec:more}.

\subsection{Parameter Analysis}
In the proposed AdaSwitch framework, two hyperparameters, $ K $ and $ L $, play pivotal roles: $ K $ regulates the threshold for switching between exploration and guidance modes, while $ L $ determines the window size for this switching mechanism.

Regarding $ K $, empirical results indicate that optimal performance is achieved when $ K $ lies between 3 and 4. Increasing $ K $ tends to elevate the proportion of exploration relative to guidance, whereas decreasing $ K $ has the opposite effect. When comparing performance across datasets, we observe that the impact of varying $ K $ is relatively minor and more stable on the SUMM dataset. In contrast, on the GSM dataset, particularly for the Llama model, performance is significantly more sensitive to changes in $ K $.

For the parameter $ L $, the best results are generally obtained around a value of 10. Larger values of $ L $ increase the minimum duration of exploration (i.e., at least $ L $ tokens), while simultaneously reducing the sensitivity of the switching mechanism to fluctuations in the KLD. Conversely, smaller $ L $ values increase sensitivity to KLD changes. Consistent with observations for $ K $, the effects of varying $ L $ are more stable and less pronounced on the SUMM dataset, whereas on the GSM dataset--and especially with the Llama model--performance exhibits greater susceptibility to $ L $ variations.
\begin{figure}[t]
    \centering
    \begin{subfigure}[b]{0.47\linewidth}
        \centering
        \includegraphics[width=\linewidth]{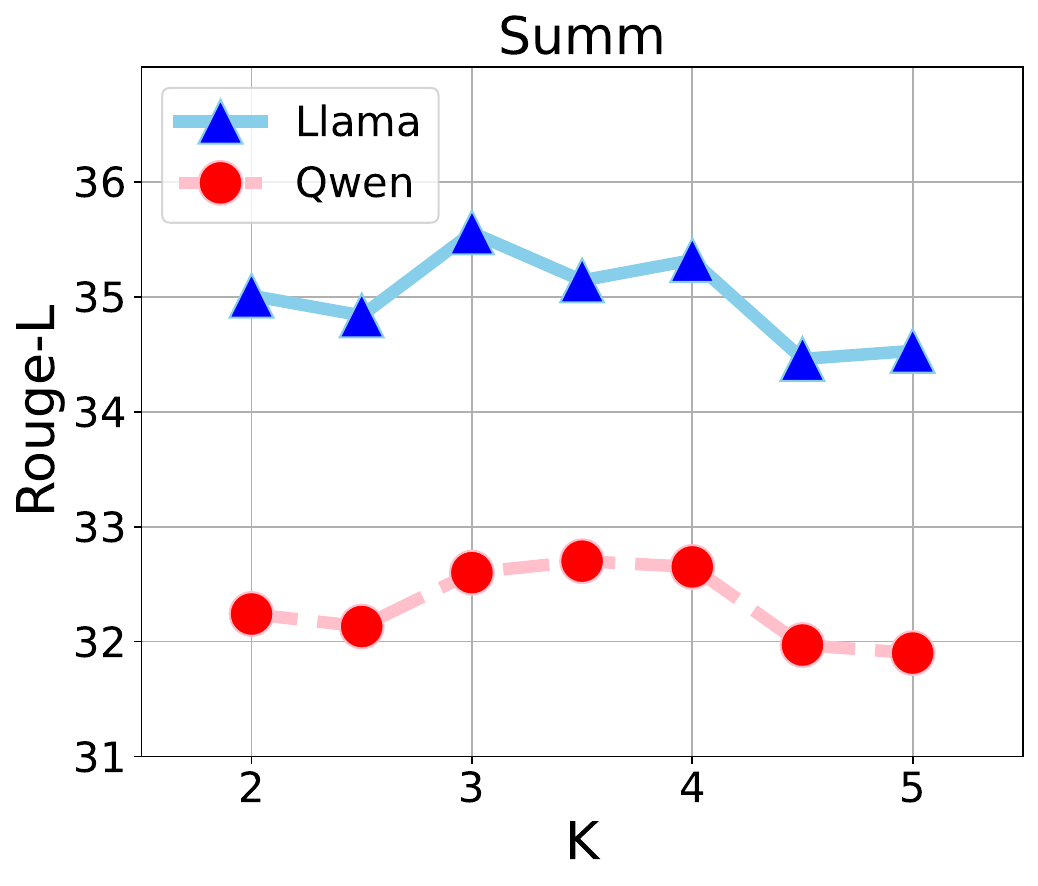}
 
    \end{subfigure}
    \hfill
    \begin{subfigure}[b]{0.47\linewidth}
        \centering
        \includegraphics[width=\linewidth]{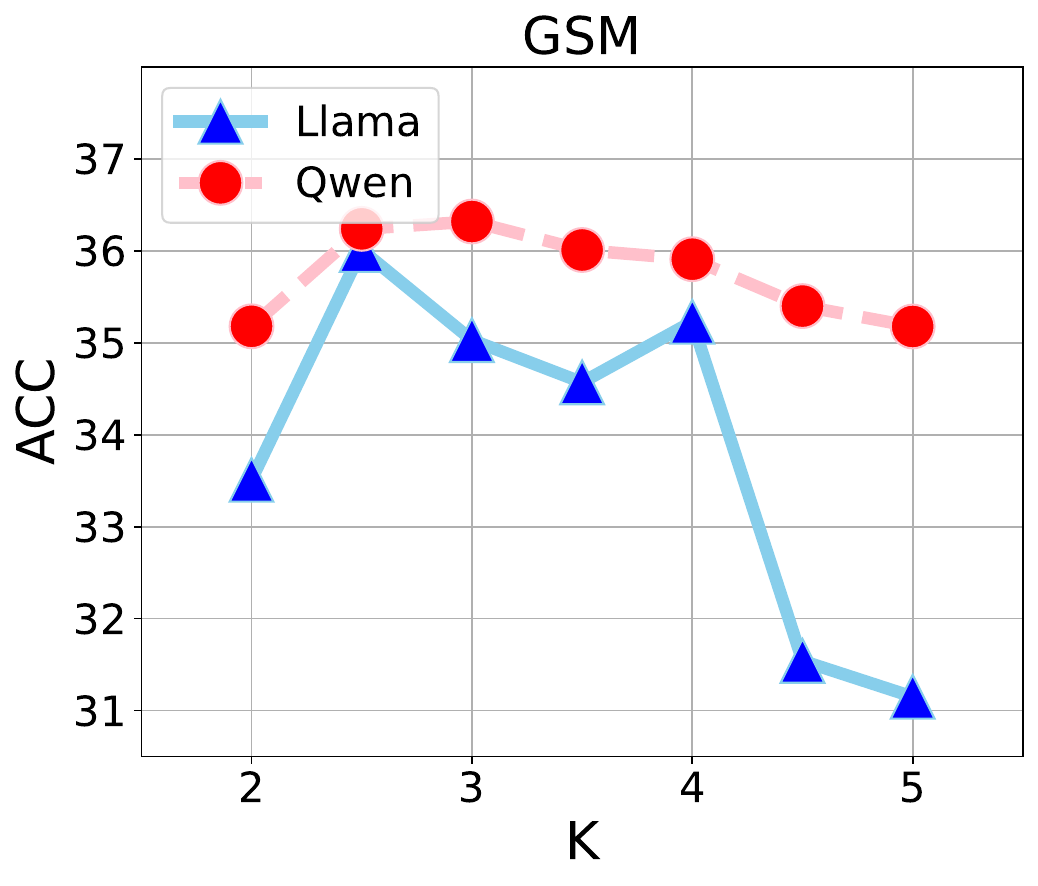}

    \end{subfigure}
    \hfill
    \begin{subfigure}[b]{0.47\linewidth}
        \centering
        \includegraphics[width=\linewidth]{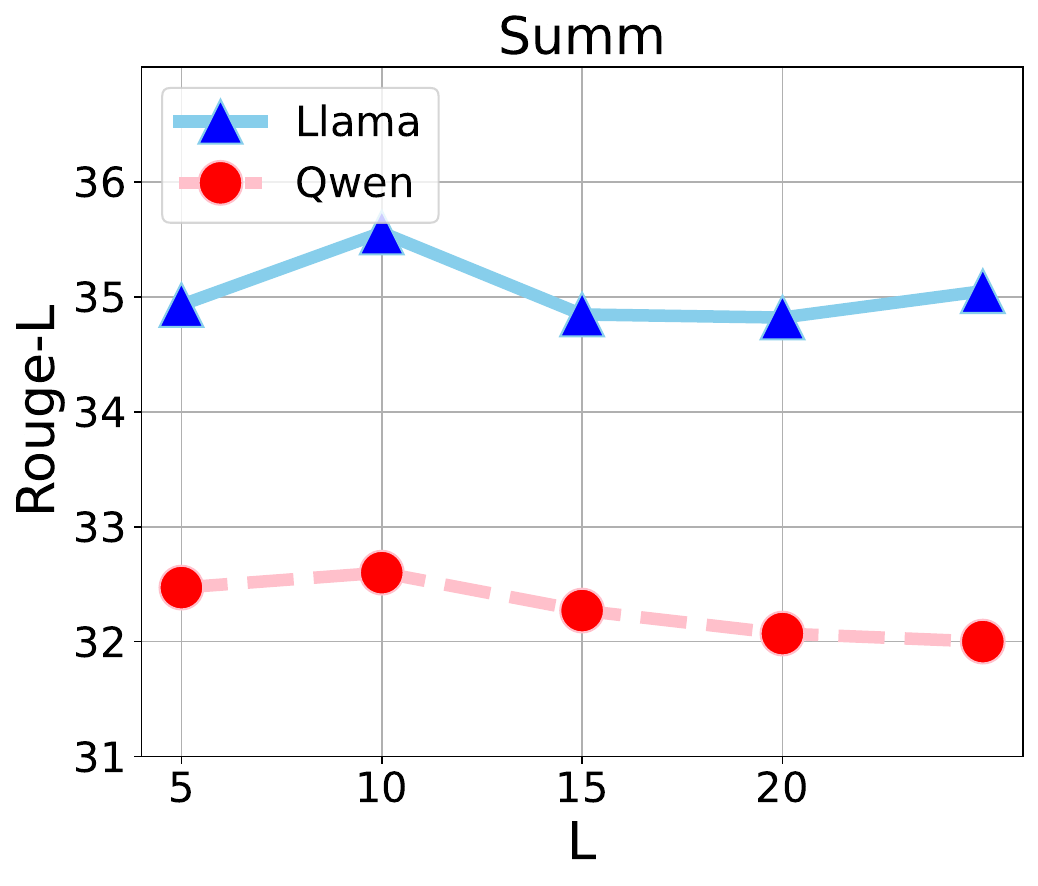}
 
    \end{subfigure}
    \hfill
    \begin{subfigure}[b]{0.47\linewidth}
        \centering
        \includegraphics[width=\linewidth]{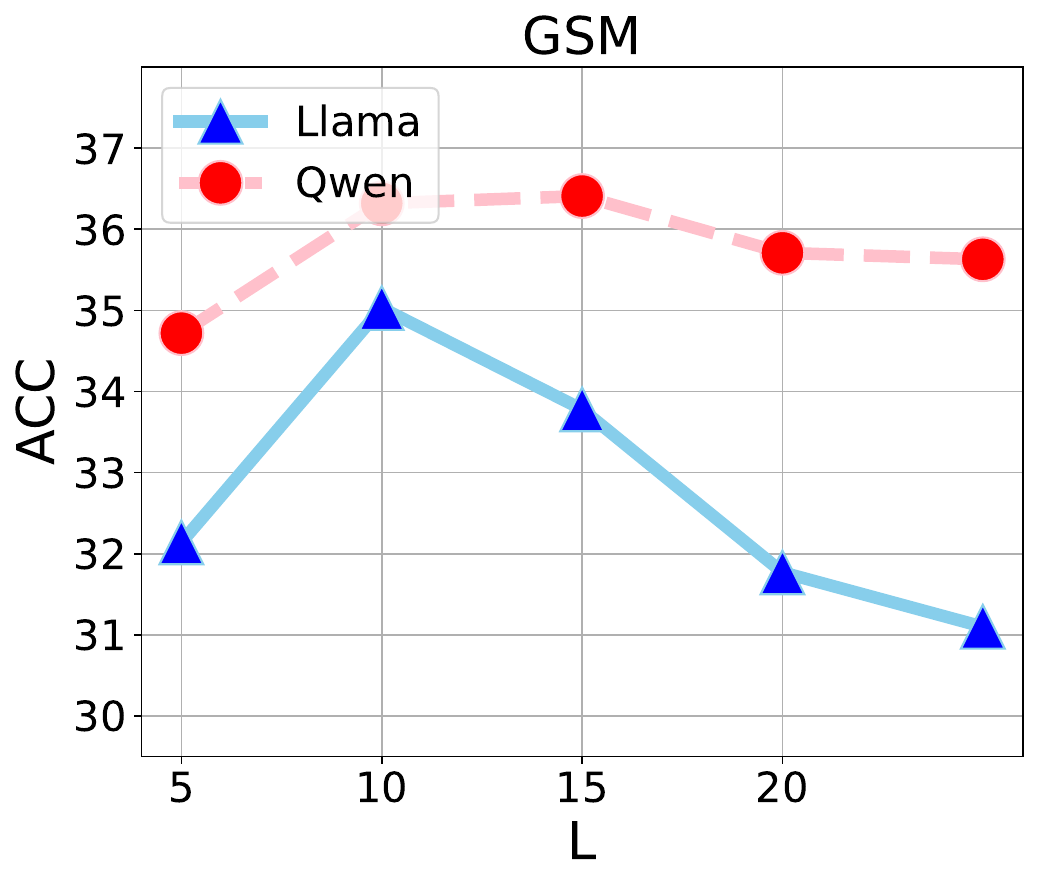}

    \end{subfigure}
    
    \caption{Parameter analysis on $K$ and $L$.}
    \label{fig:parameter}
    \vspace{-0.5cm}
\end{figure}

 \subsection{Impact of Maximum Switching Frequency}
 \label{sec:fre}
\begin{figure}[t]
    \centering
    \includegraphics[width= 0.95\linewidth]{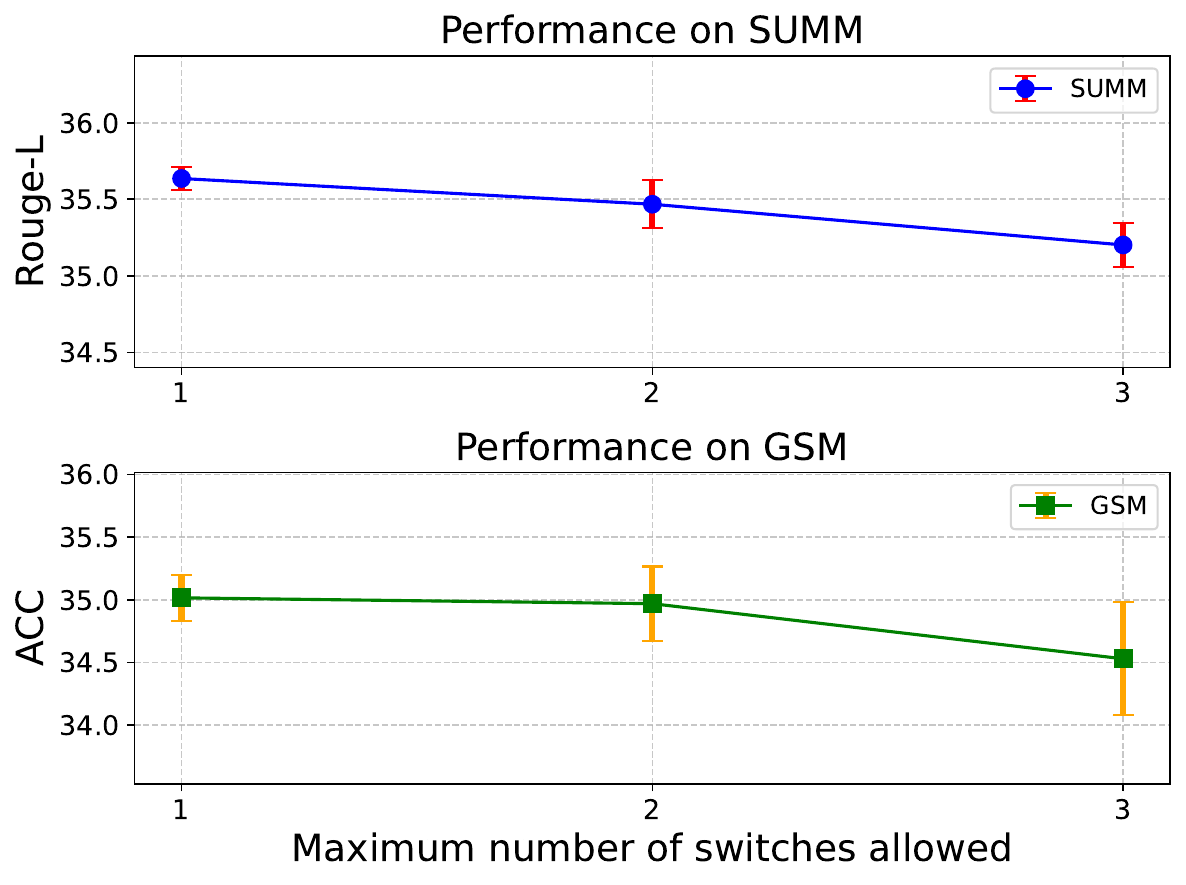}
    \caption{Performance under maximum number of switches allowed}
    \label{fig:num}
    \vspace{-0.7cm}
\end{figure}

 To validate our single-switch design, we conducted an ablation study varying the maximum switching budget from 1 to 3 on SUMM and GSM tasks using the Llama model pair. Averaged over five random seeds, the results in Figure \ref{fig:num} show that increasing the switch limit consistently degrades performance: Rouge-L on SUMM drops from 35.6 to 35.2, while GSM accuracy declines from 35.0 to 34.6. Beyond performance loss, we observe marked training instability, with error bars expanding significantly as the budget grows, particularly on GSM. This suggests that frequent student–teacher alternations introduce optimization noise and fragment the generation process into semantically inconsistent ``chimera'' sequences. These findings empirically justify AdaSwitch's one-time switching mechanism, which triggers at most once per sequence precisely at the student's capability limit, thereby preserving generation coherence while ensuring timely guidance.
 

\subsection{Time Consumption Analysis}

AdaSwitch runtime depends on sequence length and generation speed. However, estimating length across KD methods is difficult, and generation constitutes a small fraction of total time. Therefore, we compared total distillation runtime. Table~\ref{tab:time} shows off-policy is fastest as it avoids generation. Mixing-based methods are slower than on-policy due to teacher logits and switching overhead. Among them, AdaSwitch achieves the lowest overhead (1.2$\times$ on SUMM, 1.4$\times$ on GSM). This confirms that the single-switch mechanism improves efficiency by eliminating repeated alternations and redundant forward passes.

%% file: 4RelatedWork.tex
\section{Related Work}
\subsection{Small Language Model Optimization}
The scaling law suggests that increasing the number of parameters makes a language model more powerful. However, relatively small language models (SLMs) are essential in scenarios that are sensitive to latency or computational costs, such as search engines \cite{li2025towards} and recommendation systems \cite{deldjoo2024review}. Therefore, developing a small language model with satisfactory performance for specific tasks becomes crucial.

Various optimization methods can be employed to create an SLM, including Supervised Fine-Tuning (SFT), Direct Policy Optimization (DPO) \cite{rafailov2023direct}, and Knowledge Distillation (KD) \cite{DBLP:journals/corr/HintonVD15}. However, both SFT and DPO require large amounts of high-quality labeled data \cite{raghavendra-etal-2025-balancing,wang-etal-2024-self-training}, which can be a significant limitation. In contrast, KD can leverage data of arbitrary size by utilizing a teacher model, making it more scalable. Moreover, while SFT and DPO generally rely on hard labels, KD has the advantage of using intermediate representations or logits from the teacher model as supervisory signals, providing richer information for training the student model \cite{kim-etal-2022-understanding}. Given the advantages and widespread use of KD for training SLMs \cite{muralidharan2024compact,guo2025deepseek,yang2025qwen3,abdin2024phi}, consequently, our work builds on KD to refine its core mechanism for improved performance.

\begin{table}[t]
\centering

\resizebox{0.9\linewidth}{!}{
\begin{tabular}{ccccc}
\hline \hline
          & \multicolumn{2}{c}{SUMM} & \multicolumn{2}{c}{GSM} \\
          & Llama       & Qwen       & Llama      & Qwen       \\ \hline
Off-policy    & 0.77x       & 0.79x      & 0.61x      & 0.57x      \\
On-policy & 1.00x          & 1.00x         & 1.00x         & 1.00x         \\
SKD       & 1.28x       & 1.31x      & 1.61x      & 1.53x      \\
SWITCH       & 1.29x       & 1.30x      & 1.59x      & 1.55x      \\
AdaSwitch & 1.20x       & 1.22x      & 1.41x      & 1.38x      \\ \hline \hline
\end{tabular}
}
\caption{Relative runtime cost of different KD methods over the distillation process. Results are normalized by On-policy (set as 1.00x)}
\label{tab:time}
\vspace{-0.5cm}
\end{table}

\subsection{Knowledge Distillation for LLMs}
In the era of LLM, KD has been adapted to autoregressive models and categorized into off-policy and on-policy methods, inspired by reinforcement learning \cite{xu2024survey}. Off-policy KD uses either ground-truth sequences or teacher-generated logits for supervision \cite{kim2016sequence,sanh2019distilbert}. While this ensures high-quality training data, it often causes a mismatch between the student’s training distribution and inference-time behavior, leading to inconsistencies \cite{agarwal2024policy}. Conversely, on-policy KD trains the student using its own generated sequences, improving consistency and performance on some tasks. However, the student’s initially poor outputs limit sequence quality and thus KD effectiveness \cite{xuspeculative}. Prior attempts to combine off- and on-policy KD typically mix data at a coarse level \cite{lin2020autoregressive,agarwal2024policy}. Among the two most related approaches, SKD employs speculative decoding guided by teacher supervision to enhance sequence quality \cite{xuspeculative}. SWITCH \cite{koo-etal-2025-switch} alternates between the student and the teacher during sequence generation based on a threshold with a fixed decay rate. However, both methods rely on frequent switching between the student and teacher models during generation, which can lead to overfitting to the teacher and limit performance improvements. In contrast, AdaSwitch mitigates this issue by introducing a more stable and efficient token-level mixing strategy.

%% file: 5Conclusion.tex
\section{Conclusion}
In this paper, we introduce AdaSwitch, a token-level mixed distillation framework designed to reduce the gap between training and inference while preserving high-quality student outputs. AdaSwitch adopts a two-stage generation process that dynamically switches from student exploration to teacher guidance. Extensive experiments on three tasks, three LLM families, and three distance metrics verify its effectiveness, delivering significant performance gains with acceptable time cost.

%% file: 6Limitations.tex
\section{Limitations}
Our primary limitation lies in the constrained scope of our experimental evaluation. We validate AdaSwitch on three tasks and three LLM families, which may not fully capture the breadth of challenges faced in diverse real-world applications. Future work will extend the evaluation to a wider range of tasks and domains to further verify the generality and robustness of the proposed method. Additionally, while AdaSwitch achieves significant performance gains, it introduces an additional computational overhead during the KD process. As shown in our time complexity analysis, AdaSwitch incurs approximately 30\% higher time consumption compared to the standard on-policy KD method. However, we consider this trade-off acceptable because the KD process is typically performed offline, and crucially, our method introduces no additional latency or computational cost during the inference phase for the final student model.

%% file: 7Appendix.tex
\appendix
\clearpage

\begin{figure}[t]
    \centering
    \begin{subfigure}[b]{0.9\linewidth}
        \centering
        \includegraphics[width=\linewidth]{figure/switch_llama.pdf}
    \end{subfigure}
    \hfill
    \begin{subfigure}[b]{0.9\linewidth}
        \centering
        \includegraphics[width=\linewidth]{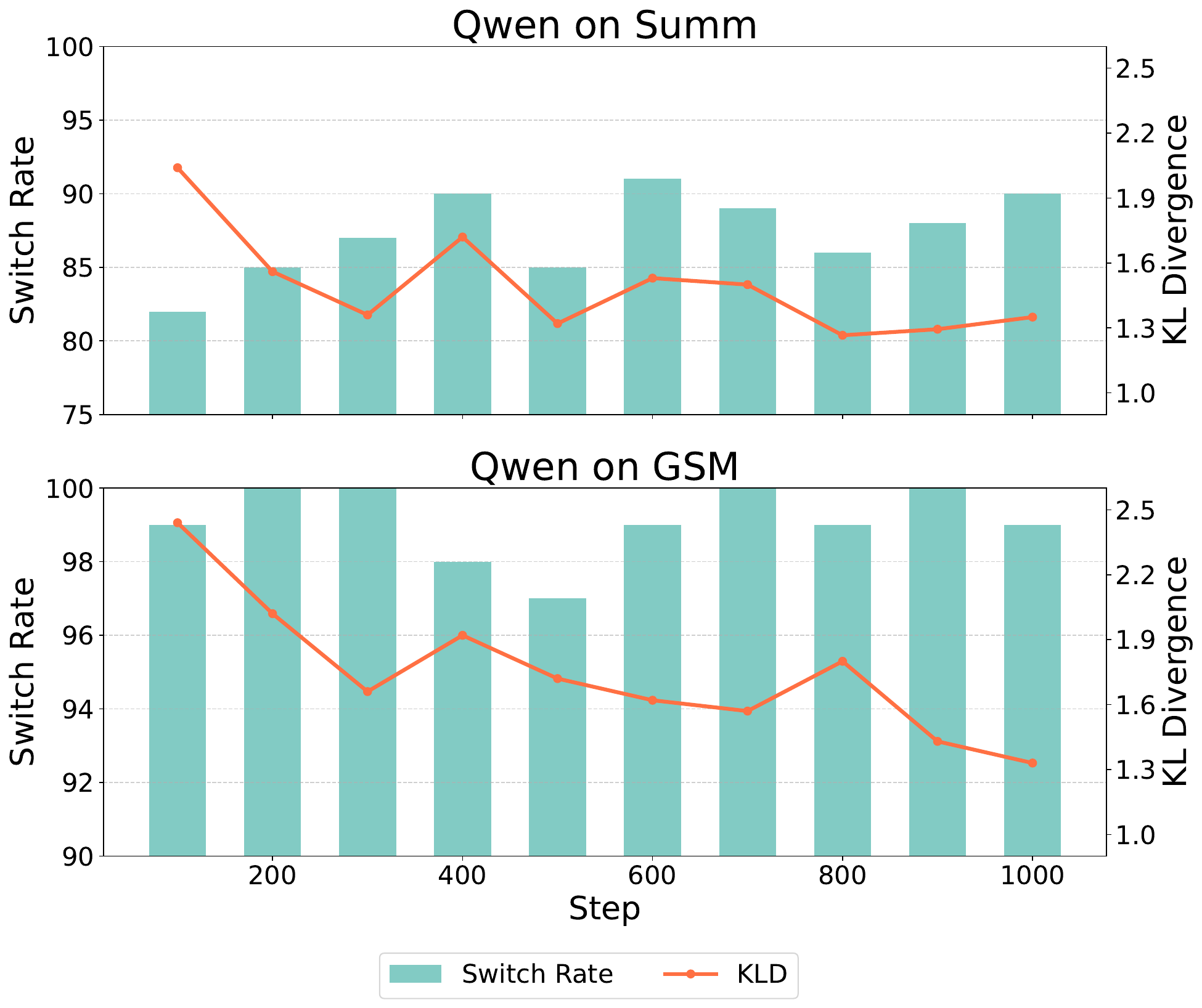}
    \end{subfigure}
    
    \caption{Analysis of the switch rate and KLD between the student and teacher models at the switch token throughout the distillation process. Statistics were collected every 100 steps for the first 1000 steps.}
    \label{fig:switch_app}
\end{figure}

\section{Datasets}
\label{sec:datasets_app}
\textbf{Dialogue Summarization.} \cite{chen2021dialogsum} For the dialogue summarization task, we employ the DialogSum dataset, which contains 12.5K training samples, 500 validation samples, and 1.5K test samples. The complete training set is utilized for SFT of the teacher models. For KD, we randomly sample 1,000 instances from the training data, and the effectiveness of the KD methods is subsequently evaluated on the held-out test set. To assess summary quality, we use the ROUGE-L metric.

\textbf{GSM8K.} \cite{cobbe2021training} To evaluate arithmetic reasoning, we utilized the GSM8K dataset, a common benchmark for assessing how well language models solve math problems. The dataset comprises 8,500 training and 1,300 test examples, with each example presenting a grade-school word problem alongside its step-by-step solution and final answer. We partitioned the GSM8K training set into a training subset (7,000 instances) for SFT of the teacher models and a validation subset (473 instances) following ~\cite{xuspeculative}. For KD, we randomly selected 1,000 problems from the training data and then evaluated the effectiveness of our KD methods on the separate test set. Accuracy is applied to assess the performance.

\textbf{GSM-Plus.} \cite{li2024gsm} GSM-Plus is an adversarial dataset that evaluates the robustness of large language models on mathematical reasoning. This challenging benchmark modifies the original GSM8K problems with numerical, arithmetic, and textual perturbations to test if models can maintain correct reasoning despite subtle changes. In our experiments, we evaluated the student models, which were distilled on GSM8K, on the GSM-Plus test set to further assess the generalizability and robustness of our knowledge distillation methods. The model's performance is assessed using accuracy.

\begin{figure}[t]
    \centering
    \begin{subfigure}[b]{0.9\linewidth}
        \centering
        \includegraphics[width=\linewidth]{figure/vali_test_llama.pdf}
    \end{subfigure}
    \hfill
    \begin{subfigure}[b]{0.9\linewidth}
        \centering
        \includegraphics[width=\linewidth]{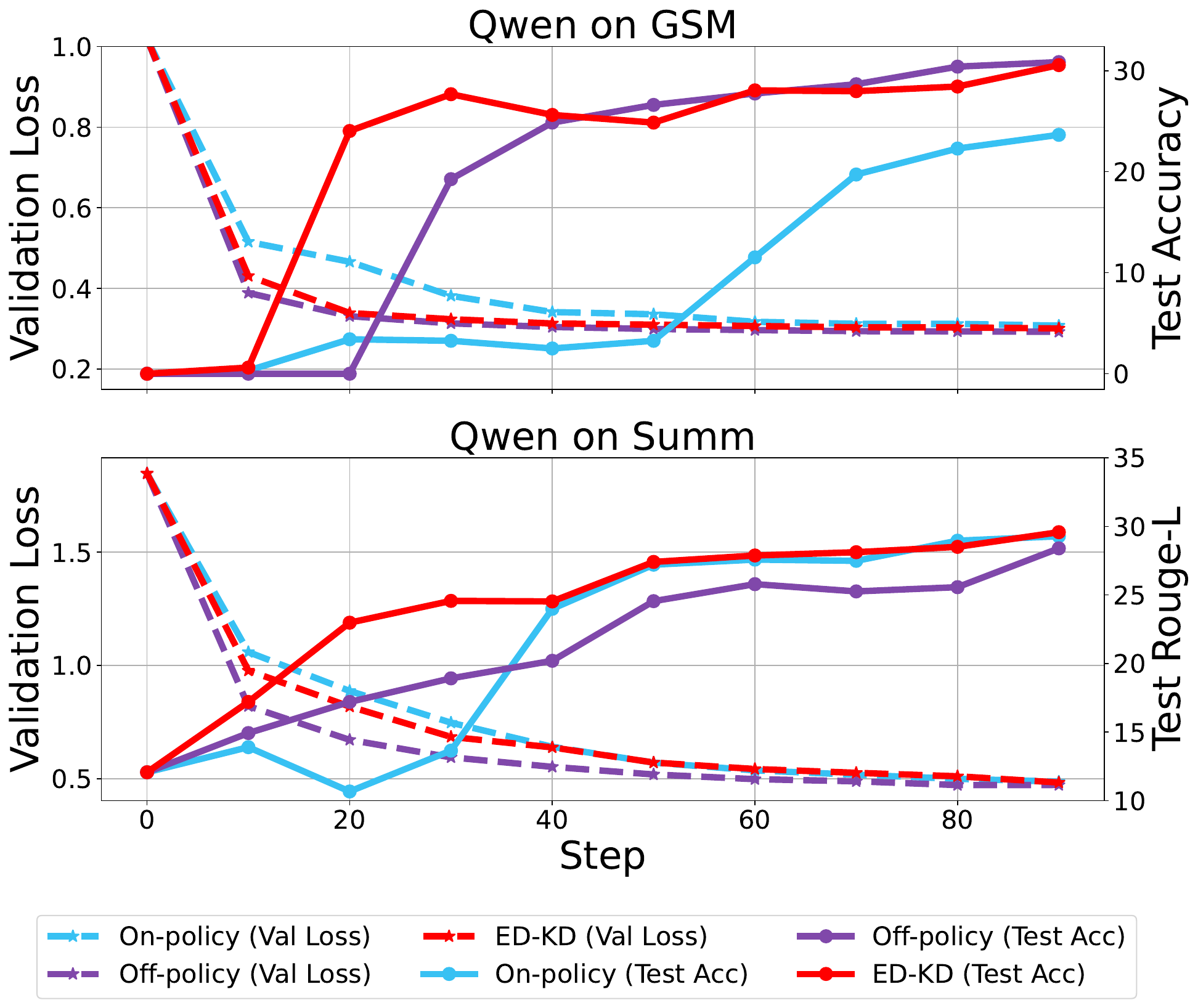}
    \end{subfigure}

    \caption{Analysis of the validation loss in early stage of distillation}
    \label{fig:vali_app}
\end{figure}

\subsection{Baselines} 
\label{sec:base}
\textbf{Supervised FT} directly minimizes the negative log-likelihood of the student’s predictions on the target sequence, which is the ground-truth.

\noindent \textbf{SeqKD} \cite{kim2016sequence} performs Supervised FT on teacher-generated target sequences.

\noindent \textbf{SupervisedKD} \cite{sanh2019distilbert} trains a student model to mimic the token-level probability distribution of a teacher model over a fixed ground-truth sequence of each $(x,y)$ pair by minimizing the distance between the student’s and teacher’s predictions. Since this method delivers the best performance among all off-policy approaches, we refer to it simply as \textbf{off-policy} in the following.

\noindent \textbf{GKD} \cite{agarwal2024policy} applied on-policy KD to address the training-inference mismatch in student models by sampling target tokens directly from the student’s own output. We refer to this method as \textbf{on-policy} in the remainder of this paper.

\noindent \textbf{ImitKD} \cite{lin2020autoregressive} randomly sample target sequences from either the ground truth
dataset or the student model, which is a sequence-level mixing strategy of on-policy and off-policy. 

\noindent \textbf{SWITCH} \cite{koo-etal-2025-switch} employs token-level switching between the student and teacher based on a fixed threshold decay during sequence generation.

\noindent \textbf{SKD} \cite{xuspeculative} employs a speculative decoding method to improve the quality of the generated sequences by using the teacher model to supervise the sequence generation.

\section{More Experimental Results}
\label{sec:more}

We provide the complete experimental results for the Analysis of Switch Rate and KLD, as well as the Early-Stage Distillation Analysis, in Figure~\ref{fig:switch_app} and Figure~\ref{fig:vali_app}, respectively.